\pdfoutput=1

\documentclass[11pt]{article}

\usepackage[]{EMNLP2023}

\usepackage{amsmath,amsfonts,bm}

\def\eqref#1{equation~\ref{#1}}

\def\1{\bm{1}}

\DeclareMathAlphabet{\mathsfit}{\encodingdefault}{\sfdefault}{m}{sl}
\SetMathAlphabet{\mathsfit}{bold}{\encodingdefault}{\sfdefault}{bx}{n}

\usepackage{wrapfig,lipsum,booktabs}
\usepackage{hyperref}
\usepackage{url}
\usepackage{graphicx}
\usepackage{amsmath,amssymb,amsfonts}
\usepackage{pifont}
\usepackage{array}
\usepackage{multirow}
\usepackage[export]{adjustbox}
\usepackage{caption}
\usepackage{subcaption}
\usepackage{float}
\usepackage{newfloat}
\usepackage{makecell}
\usepackage{bbm}
\usepackage{bm}	
\usepackage{subcaption}
\usepackage{placeins}
\usepackage{listings}
\usepackage{bibentry}
\usepackage{arydshln}
\usepackage[noend]{algpseudocode}

\usepackage{times}
\usepackage{latexsym}

\usepackage[T1]{fontenc}

\usepackage[utf8]{inputenc}

\usepackage{microtype}

\usepackage{inconsolata}

\title{MerA: Merging Pretrained Adapters For Few-Shot Learning}

\author{Shwai He\textsuperscript{\rm 1}\space\space
Run-Ze Fan\textsuperscript{\rm 3}\space\space
Liang Ding\textsuperscript{\rm 2}\thanks{~~Corresponding author}\space\space
Li Shen\textsuperscript{\rm 4}\space\space 
Tianyi Zhou\textsuperscript{\rm 1}\space\space 
Dacheng Tao\textsuperscript{\rm 2}\\
    \textsuperscript{\rm 1}University of Maryland, College Park \space\space
    \textsuperscript{\rm 2}The University of Sydney\\
    \textsuperscript{\rm 3}University of Chinese Academy of Sciences \space\space
    \textsuperscript{\rm 4}JD Explore Academy\\
    {\tt\small shwaihe@umd.edu},\space\space
    {\tt\small liangding.liam@gmail.com}\space\space
}

\makeatletter
\newcommand{\biggg}{\bBigg@{1.2}}

\makeatother

\newcommand{\ignore}[1]{{}}

\begin{document}

\maketitle

\begin{abstract}

Adapter tuning, which updates only a few parameters, has become a mainstream method for fine-tuning pretrained language models to downstream tasks. However, it often yields subpar results in few-shot learning. AdapterFusion, which assembles pretrained adapters using composition layers tailored to specific tasks, is a possible solution but significantly increases trainable parameters and deployment costs. Despite this, our preliminary study reveals that even single adapters can outperform Adapterfusion in few-shot learning, urging us to propose \textbf{\texttt{Merging Pretrained Adapters}} (MerA) that efficiently incorporates pretrained adapters to a single model through model fusion. Extensive experiments on two PLMs demonstrate that MerA achieves substantial improvements compared to both single adapters and AdapterFusion. To further enhance the capacity of MerA, we also introduce a simple yet effective technique, referred to as the "\textit{same-track}" setting, that merges adapters from the same track of pretraining tasks. With the implementation of the "\textit{same-track}" setting, we observe even more impressive gains, surpassing the performance of both full fine-tuning and adapter tuning by a substantial margin, e.g., 3.5\% in MRPC and 5.0\% in MNLI. 
\end{abstract}



\section{Introduction}
\label{sec:introduction}

Pretrained language models (PLMs) \cite{devlin2018bert, liu2019roberta} have revolutionized the field of natural language processing, with fine-tuning being a mainstream approach to leverage the power of PLMs. However, with the ever-increasing number of parameters in PLMs \cite{NEURIPS2020_1457c0d6}, there is a need for parameter-efficient fine-tuning techniques \cite{DBLP:conf/iclr/HeZMBN22} to reduce training costs. One representative technique is adapters tuning \cite{houlsby2019parameter}, which updates only a subset of parameters. Despite their advantages, adapter tuning often starts with randomly initialized blocks and may not perform well in scenarios with limited training data, such as few-shot learning \cite{DBLP:conf/naacl/MoosaviDKG22, bansal2022meta}.

\begin{figure}[t]
\centering
\makeatother\def\@captype{figure}\makeatother
	\centering
	\includegraphics[width=0.48\textwidth]{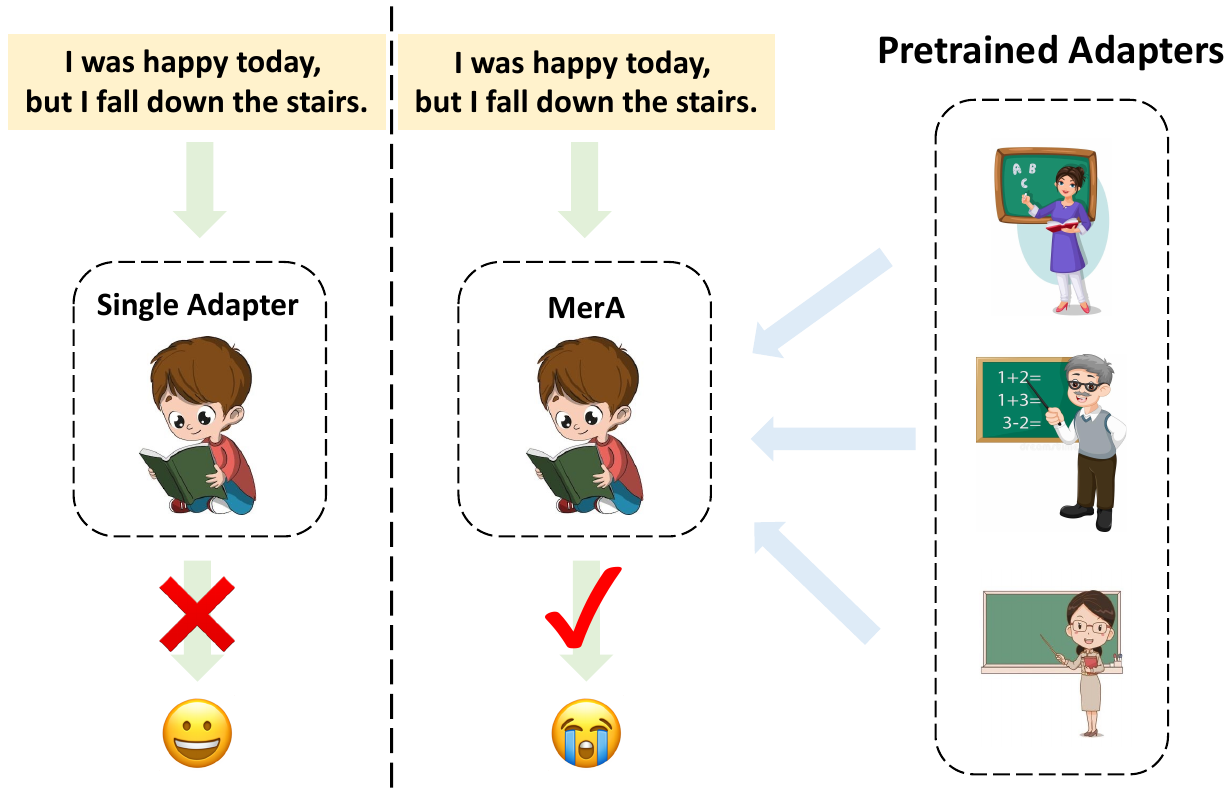}
    \caption{Comparison between a single adapter and MerA in sentiment analysis tasks. Left: The dilemma of a single adapter in \textbf{few-shot learning}. Right: the improvement of merging pretrained adapters. }
    \label{fig: poster}
\end{figure}

One possible solution is to transfer knowledge from pretrained adapters to target tasks \cite{Chawla_2021_WACV,zhong2022panda,pmlr-v180-wang22a}. AdapterFusion~\cite{pfeiffer2020adapterfusion} has been proposed to assemble pretrained adapters with composition layers to integrate knowledge. However, AdapterFusion compromises parameter efficiency~\cite{He2022SparseAdapterAE}, primarily due to the excessive number of trainable parameters in the composition layers. On the other hand, deploying parallel pretrained adapters also increases computational costs. 

In this work, we turn to explore an efficient approach to leverage pretrained adapters, raising the following question: \textit{Can the current AdapterFusion framework fully exploit the pretrained adapters under few-shot scenarios? If not, how to incorporate the pretrained adapters more efficiently?} To this end, we first conduct a series of experiments to compare the performance of single adapters and that of AdapterFusion under few-shot scenarios, with results shown in Figure \ref{pre_experiment}. Surprisingly, a single adapter outperforms AdapterFusion with much few trainable parameters. Such preliminary study prompts us to \textit{directly leverage pretrained adapters to extend the potential of single adapters}.

To achieve this, we propose an approach that merges pretrained adapters into a single one (MerA). On the one hand, the merge pretrained adapter does not introduce additional trainable parameters. On the other hand, the knowledge from pretrained adapters enhances downstream performance, as illustrated in Figure~\ref{fig: poster}. We first implement two straightforward methods for merging parameters, including summation (``Sum.'') and averaging (``Avg.''), whereas the lack of one-to-one correspondences between the parameters of different models leads to suboptimal performance. Inspired by~\citet{DBLP:journals/tog/SolomonGPCBNDG15, singh2020model}, we further propose to align adapters' parameters through optimal transport based on weights (``Wts.'') and activations (``Acts.'').

Extensive few-shot experiments demonstrate that MerA achieves significant improvements compared with Adapters, e.g., 2.7\% in averaged accuracy. In addition, we also find that merging adapters from the same track of tasks further enhances the capacity of MerA. Therefore, we introduce a simple yet effective technique called the "\textit{same-track}" setting. With the implementation of the "\textit{same-track}" setting, we observe even more impressive gains, surpassing the performance of full fine-tuning and Adapter tuning by a substantial margin, e.g., 3.5\% in MRPC and 5.0\% in MNLI.

\section{Methodology}
\label{sec:method}

\textbf{AdapterFusion. }
Adapters \cite{houlsby2019parameter} are bottleneck modules plugged in PLMs, with model dimensions $d$ and reduction factor $r$. In standard Adapter Tuning, only adapter layers are trainable, while other layers are frozen. After tuning, adapters contain specific knowledge for a single task. AdapterFusion \cite{pfeiffer2020adapterfusion} is proposed to leverage knowledge from pretrained adapters, which improves the performance in downstream tasks and prevents catastrophic forgetting for each adapter. 

\noindent\textbf{Motivation. }
However, the resource constraint scenario challenges the parameter efficiency of AdapterFusion. For one thing, AdapterFusion requires composition layers that include additional trainable parameters, e.g.,  \textit{Query}, \textit{Key}, and \textit{Value}. The trainable parameters of a composition layer are $3d^2$ (ignoring bias terms for the time being), while a single adapter only contains $\frac{2d^2}{r}$. On the other hand, AdapterFusion has to assign parallel pretrained adapters for each adapter layer, which multiplies the additional deployment cost. 

\begin{figure}[t]
\centering
\makeatother\def\@captype{figure}\makeatother
	\centering
	\includegraphics[width=0.46\textwidth]{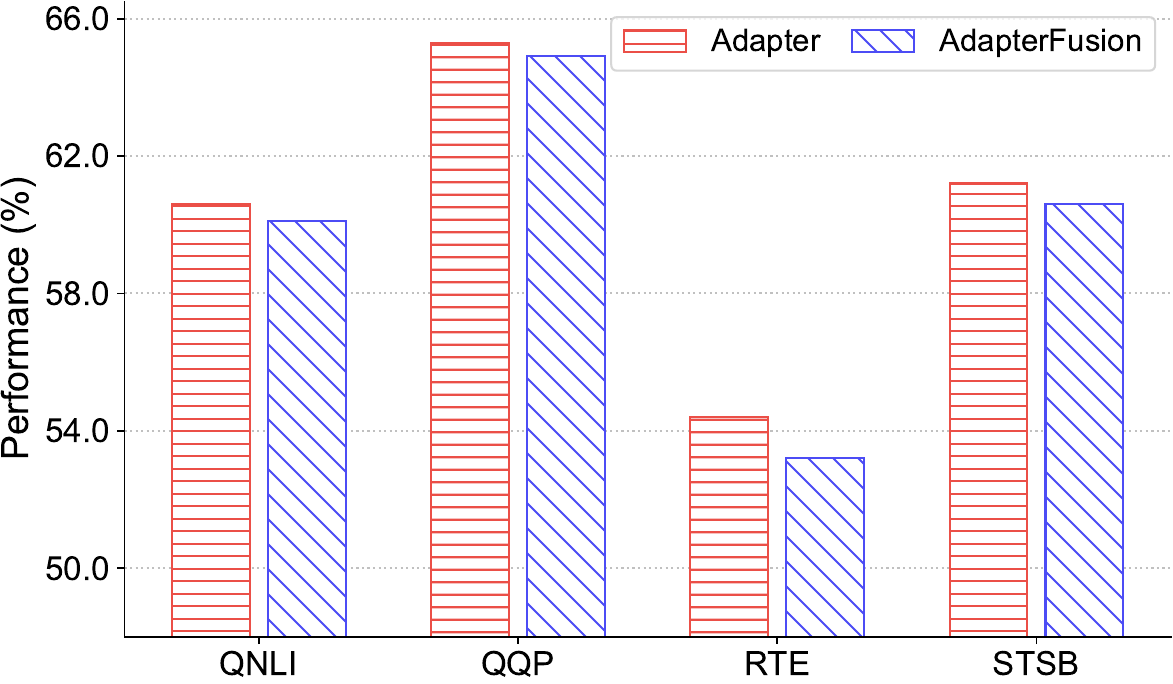}
 \caption{\textbf{Comparison between single adapters and AdapterFusion} under 10-shot scenarios. }
\label{pre_experiment}
\end{figure}

Despite the parameters budgets, to check whether AdapterFusion is an optimal choice to improve the task-specific performance under data constraint scenarios, we conduct few-shot experiments to compare single adapter and AdapterFusion. As is shown in Figure \ref{pre_experiment}, single adapters consistently outperform AdapterFusion with fewer parameters, which indicates the superiority of tuning single adapters. Inspired by the preliminary study, we intend to merge multiple adapters into a single model to explore the potential of single adapters. 

\begin{table*}[htbp]
\centering
\caption{\textbf{Experimental results of different merging Methods} on GLUE benchmark, where we perform merging with the same pretrained adapters for a fair comparison. All tasks are evaluated using accuracy. Average scores on all shots are \underline{underlined}. The best results are \textbf{bold}. 
}
\resizebox{\linewidth}{!}{
\begin{tabular}{lcccccccccccccc}
    \toprule
    \multirow{2}{*}{\bf Method}  & \multirow{1}{*}{\#Param.} & \multicolumn{4}{c}{MRPC} & 
    \multicolumn{4}{c}{SST-2} & \multicolumn{4}{c}{MNLI} \\
    
    \cmidrule(lr){3-6} \cmidrule(lr){7-10} \cmidrule(lr){11-14} 
    & (Trained)
    & \makecell{10} & \makecell{20} & \makecell{30} & \underline{Avg.} 
    & \makecell{10} & \makecell{20} & \makecell{30} & \underline{Avg.} 
    & \makecell{10} & \makecell{20} & \makecell{30} & \underline{Avg.}  \\
    
    \midrule
    Fine-Tune & 100\%
    & 66.2 & 67.6 & 69.1 & \underline{67.6} & 60.7 & 62.3 & 63.9 & \underline{62.3} & 39.4 & 41.1 & 42.2 & \underline{40.9} \\
    \midrule
    AdapterFusion & 18\%
    & 65.2 & 66.4 & 67.5 & \underline{66.4} & 60.1 & 61.5 & 63.6 & \underline{61.7} & 39.3 & 40.8 & 41.7 & \underline{40.6} \\
    Adapter & 0.8\%
    & 65.4 & 66.8 & 67.7  & \underline{66.6} & 60.2 & 61.1 & 64.1 & \underline{61.8} & 39.1 & 41.0 & 42.0 & \underline{40.7} \\
    \hdashline
    ~~w/ Sum & 0.8\%
    & 67.4 & 67.9 & 68.6 &  \underline{68.0} & 61.2 & 62.1 & 64.2 & \underline{62.5} & 39.2 & 41.4 & 42.0 & \underline{40.9} \\
    ~~w/ Avg & 0.8\%
    & 67.2 & 67.7 & 68.4 &  \underline{67.8} & 60.6 & 62.6 & 63.7 & \underline{62.3} & 39.4 & 41.7 & 42.1 & \underline{41.1} \\
    ~~w/ Wts & 0.8\%
    & 67.4 & 68.4 & 69.6 &  \underline{68.5} & 60.7 & \bf 63.3 & 65.4 & \underline{63.1} & \bf 39.9 & 41.8 & 42.4 & \underline{41.4} \\
    ~~w/ Acts & 0.8\%
    & \bf 67.6 & \bf 68.8 & \bf 70.3 &  \underline{\bf 68.9} & \bf 61.6 & 63.2 & \bf 65.6 & \underline{\bf 63.5} & 39.7 & \bf 42.1 & \bf 42.6 & \underline{\bf 41.5} \\
    \bottomrule
    \end{tabular}}
\label{tab:main_results}
\end{table*}
\noindent\textbf{Adapter Merging. }
Due to the limitations of AdapterFusion, we intend to make more efficient use of the knowledge in the pretrained adapters.  Integrating the parallel adapter into a module can reduce the excess trainable parameters and improve downstream performance.  We first consider two simple methods for merging the weights in adapters trained on different tasks.  The first one is the summation (``Sum.''): 
\begin{equation}
\small
   \widetilde W = \sum\nolimits_{j=1}^n W_{\tau_j}, 
\end{equation}
where we denote $\tau$ as the indices of the tasks and weights of the adapter trained on task $\tau_j$ as $W_{\tau_j}$. The second one is averaging (``Avg.''): 
\begin{equation}
\small
   \widetilde W = \frac{1}{n} \sum\nolimits_{j=1}^n W_{\tau_j}. 
\end{equation}
However, the problem with vanilla summation and averaging is the lack of one-to-one correspondence between parameters from different adapters. In particular, the $p$-th neuron of the $i$-th adapter might behave very differently from the $p$-th neuron of the $j$-th adapter and instead behave similarly to another neuron. Therefore, aligning the neurons first and then assembling adapters makes sense. Inspired by \citet{singh2020model,zan-etal-2022-complementarity}, we align adapters' parameters via optimal transport based on weights (``Wts.'') and activations (``Acts.''). 

Given adapters trained from task $\tau_i$ and $\tau_j$, we plug them into language models and denote the $l$-th adapter layer's incoming edges as $W^{(\ell,\, \ell - 1)}_{\tau_i}$ and $W^{(\ell,\, \ell - 1)}_{\tau_j}$, respectively. We align $W^{(\ell,\, \ell-1)}_{\tau_j}$ to $W^{(\ell,\, \ell-1)}_{\tau_i}$ by constructing convex combinations of previous layer transport matrix $T^{(\ell - 1)}$ based on weights (``Wts'') or activations (``Acts''), normalized appropriately via the inverse of corresponding column marginals $\beta$: 
\begin{equation} 
\small
\widehat{W}^{(\ell, \,\ell-1)}_{\tau_j}   \leftarrow W^{(\ell,\, \ell-1)}_{\tau_j} T^{(\ell - 1)} \text{diag} (1 / \beta^{(\ell - 1)}), 
\end{equation}
\begin{equation} 
\small
\widetilde{W}^{(\ell, \,\ell-1)}_{\tau_j}   \leftarrow \text{diag} (1 / {\beta^{(\ell)}}) {T^{(\ell)}}^{\top} \widehat{W}^{(\ell,\, \ell-1)}_{\tau_j}. 
\end{equation}
We refer $\widetilde{W}^{(\ell, \,\ell-1)}_{\tau_j}$ to the aligned weights of the adapter from task $\tau_j$, which can be directly added with $W^{(\ell, \,\ell-1)}_{\tau_i}$. We carry out this procedure on all pretrained adapters: 
\begin{equation}
\small
\widetilde{W}^{(\ell, \,\ell - 1)}   \leftarrow 
\frac{1}{n}\sum\nolimits_{j = 1}^n {W}_{r_j}^{(\ell, \,\ell-1)}. 
\end{equation}

\section{Empirical Evaluation}
\label{sec:experiments}

\subsection{Setup}
We collect pretrained adapters from AdapterHub \cite{pfeiffer2020AdapterHub}, which are trained on imdb \cite{maas-EtAl:2011:ACL-HLT2011}, boolq \cite{clark2019boolq}, scitail \cite{khot2018scitail} and winogrande \cite{sakaguchi2021winogrande}. The pretraining tasks cover different NLP tracks, including sentiment analysis, question-answering, natural language inference, and common sense reasoning. Our experiments were conducted on the widely-used GLUE benchmark~\cite{wang2018glue}. 

We use Adam~\cite{kingma2014adam} as the optimizer with $\beta_1$, $\beta_2$ = 0.9, 0.98. We set the weight decay as 0.1 and grid-search the learning rate and training epochs from \{1e-5, 5e-5, 1e-4\}, and \{5, 10\}. The maximum length is 128. We follow previous works~\cite{phang2018sentence, lee2019mixout, dodge2020fine,he-etal-2023-pad,zhong-etal-2023-revisiting} to fine-tune the pretrained language models, e.g., BERT~\cite{devlin2018bert}, on the downstream training set and report results on the dev set using the last checkpoint. 

\begin{table*}[htbp]
\centering
\caption{\textbf{Comparison between MerA and adapter architectures under various fine-tuning strategies.} We display the performance of 30 shots tuning and denote head-based tuning, prompt-based tuning, and prompt-based tuning with demonstrations as ``head'', ``prompt'', and ``prompt-demo'', respectively.}
\resizebox{\linewidth}{!}{
\begin{tabular}{lccccccccc}
    \toprule
    \multirow{2}{*}{\bf Method}  & \multirow{1}{*}{\#Param.} & 
    \multicolumn{4}{c}{BERT-Base} & 
    \multicolumn{4}{c}{RoBERTa-Base} \\
    
    \cmidrule(lr){3-6} \cmidrule(lr){7-10} & (Trained)
    & \makecell{head} & \makecell{prompt} & \makecell{+demos} & \underline{Avg.} 
    & \makecell{head} & \makecell{prompt} & \makecell{+demos} & \underline{Avg.} \\
    \midrule
    Fine-Tune & 100\%
    & 63.9 & 71.8 & 72.6 & \underline{69.4} & 71.1 & 79.4	& 79.9 & \underline{76.8}\\
    \midrule
    AdapterFusion & 18\%
    & 63.6 & 71.2 & 71.7 &  \underline{68.8} & 70.9 & 78.9 & 79.4 & \underline{76.1} \\
    Adapter & 0.8\%
    & 64.1 & 71.5 & 72.1  & \underline{69.2} & 70.5	& 78.5 & 79.3 & \underline{76.4}\\
    MerA & 0.8\%
    & \textbf{65.7}	& \textbf{73.6} & \textbf{73.9} & \underline{\textbf{71.1}} & \textbf{73.3} & \textbf{81.3} & \textbf{82.1} & \underline{\textbf{78.9}}
    \\\bottomrule
       \end{tabular}}
\label{tab:prompt-based-setting}
\end{table*}

\subsection{Results}
\paragraph{Main Results.}
In Table~\ref{tab:main_results}, we carefully compare MerA (with merging methods above: ``Sum.'', ``Avg.'', ``Wts'', ``Acts'') to the standard adapter~\cite{houlsby2019parameter} (``Adapter'') and AdapterFusion \cite{pfeiffer2020adapterfusion} (``AF'') on GLUE benchmark for pretrained language models BERT \cite{devlin2018bert}, where we set training shots to 10, 20, 30, respectively.  

MerA achieves significantly better performance than vanilla adapter~\cite{houlsby2019parameter, pfeiffer2020adapterfusion} with the same parameters budget. Compared to AdapterFusion \cite{pfeiffer2020adapterfusion}, our methods significantly reduce required trainable parameters and improve performance. 
\begin{table}[ht]
    \centering
    \caption{\textbf{Effect of different tracks. } We denote a single pretrained adapter as ``+Single'', and denote MerA with pretrained tracks (e.g., ``+QA''). For each track, the same number of pretrained adapters are chosen for MerA. }
    \resizebox{\columnwidth}{!}{
    \setlength{\tabcolsep}{2pt}
    \begin{tabular}{lllll}
    \toprule
    \multirow{2}{*}{\bf Method}  & 
    \multicolumn{2}{c}{MRPC} & \multicolumn{2}{c}{MNLI} \\
    \cmidrule(lr){2-3} \cmidrule(lr){4-5}
    & \makecell{10} & \makecell{20} & \makecell{10} & \makecell{20} \\
    \midrule
     Adapter & 65.4 & 67.0 & 39.4 & 40.8 \\ \hdashline
     +Single & 65.0 & 67.4 & 40.6 & 42.3 \\
     +QA  & 66.9 & 68.9 & 39.2 & 40.8 \\
     +NLI & 66.4 & 68.6 & \bf $\text{44.2}^{\Uparrow +4.8}$ & \bf $\text{45.8}^{\Uparrow +5.0}$ \\
     +Sentiment & 64.2 & 66.9 & 39.8 & 41.9 \\
     +Comsense & 65.4 & 68.2 & 39.6 & 41.8 \\
     +STS   & \bf $\text{68.9}^{\Uparrow +3.5}$ & \bf $\text{70.1}^{\Uparrow +3.1}$ & 40.2 & 42.2 \\
    \bottomrule
    \end{tabular}}
\label{tab:tracks}
\end{table}

\paragraph{MerA with Different Merging Methods}
MerA from all merging methods outperforms the standard adapter tuning, where optimal transport methods outperform two naive methods in all tasks, reflecting the necessity of weight alignment. Notably, ``Acts'' based Adapters achieve up to 2.7\% average improvement compared to the standard adapter and even beat full fine-tuning, so we set ``Acts'' as the default setting in the following experiments.
\begin{figure}[t]
\centering
\makeatother\def\@captype{figure}\makeatother
	\centering
\begin{subfigure}[h]{0.495\linewidth}
    \centering
    \includegraphics[width=\linewidth]{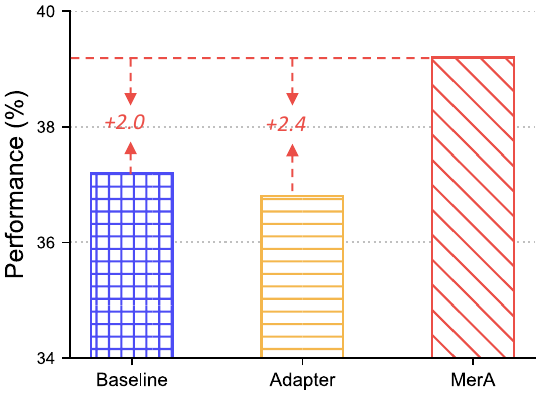}
    \subcaption{MNLI}
\end{subfigure}\begin{subfigure}[h]{0.495\linewidth}
    \centering
    \includegraphics[width=\linewidth]{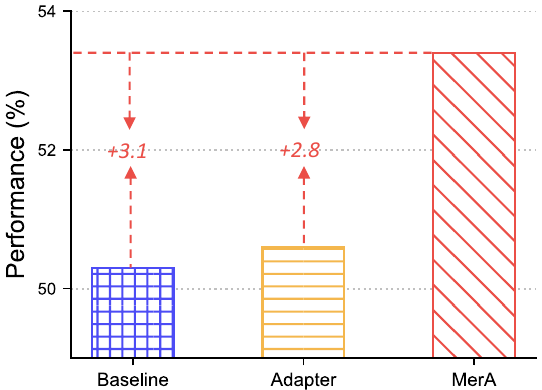}
    \subcaption{SST-2}
\end{subfigure}
\caption{\textbf{Effect of MerA on Initialization.} ``Baseline'' represent vanilla BERT. We denote BERT equipped with adapters as ``Adapters'' and denote BERT equipped with ``MerA'' as ``MerA'', respectively. } 
\label{fig:zero-shot}
\end{figure}

\paragraph{Prompt-Based Finetuning}
In addition to head-based fine-tuning, we also investigate the applicability of MerA in prompt-based fine-tuning, which leverages informative and indicative prompts to enhance performance \cite{gao-2021-making}. In Table~\ref{tab:prompt-based-setting}, we consider two strategies including prompt-based finetuning and prompt-based finetuning with demonstrations. Experimental results demonstrate the consistent improvement of MerA across various fine-tuning strategies. This highlights the effectiveness and versatility of MerA in enhancing performance in various fine-tuning scenarios. 
\paragraph{Effect on Initialization}
To analyze the benefits of merging adapters, we conducted a zero-shot experiment to examine the effect of MerA on task initialization. In Figure \ref{fig:zero-shot}, we plugged MerA into BERT and compared it with BERT models equipped with randomly initialized adapters (``Adapters'') and without any adapters (``Baseline''). Compared to ``Adapter'' and ``Baseline'', MerA ensures a superior initial state for the target tasks and achieves significant accuracy improvement, e.g., 2.4\% in MNLI and 0.98\% in SST-2. These findings reveal the efficacy of MerA in enhancing the initialization of target tasks.

\paragraph{Augmenting MerA with Same-Track Setting}
One potential approach to further increase the gain of MerA at initialization is to merge the adapters trained in the same NLP track because of the sharing of knowledge in a specific track. So we further investigate the roles of different pretrained tracks. We consider pretrained adapters from five different tracks, including question-answering (QA), common sense reasoning (Comsense), natural language inference (NLI), semantic textual similarity (STS), and sentiment analysis (Sentiment). MRPC and MNLI belong to semantic textual similarity and natural language inference, respectively. We merge adapters trained in the same track and validate MerA on MNLI and MRPC. We also consider directly fine-tuning a single pretrained adapter ``+Single'', experimenting with multiple adapters within the same tracks, and reporting the best results. 

Table \ref{tab:tracks} compares MerA merged from different tracks, where we can see significant improvements when the pretraining tasks are on the same track as the downstream task. In this case, MerA outperforms the standard adapter by 3.5\% in MRPC and 5.0\% in MNLI. However, when pretrained tasks are on a different track, MerA may encounter a performance drop. The above findings inform the importance of the knowledge shared in one track.

\section{Conclusion}
\label{sec:conclusion}
In this work, we systematically investigate adapters and AdapterFusion on few-shot scenarios. Based on our findings, we propose a plug-in strategy, i.e., MerA, for existing adapters. Our empirical results indicate the potential to make MerA a golden standard efficient few-shot learning strategy for the NLP community. 

\section{Limitations}
\label{:sec:Limitations}
Despite the progress we made, there still exist limitations in our work. On the one hand, we only investigated some classic merging methods and found that ``Acts'' performs the best in selected criteria. However, other advanced pruning methods may exist that can further improve the performance, which deserves exploration in future work. On the other hand, since we only consider BERT and RoBERTa in limited tasks, it would be valuable to consider other architecture families (e.g., XLNET \cite{yang2019xlnet}, ELECTRA \cite{clark2020electra}) and tasks (e.g., machine translation). 

\section*{Ethics Statement}
We take ethical considerations very seriously. This paper focuses on higher model and data efficiency for Adapters. Both the datasets and models used in this paper are publicly available and have been widely adopted by researchers. We ensure that the findings and conclusions of this paper are reported accurately and objectively. 

\bibliography{PaperForReview}
\bibliographystyle{acl_natbib}

\end{document}